\documentclass[11pt,a4paper]{article}
\usepackage[hyperref]{acl2019}
\usepackage{times}
\usepackage{latexsym}

\usepackage{url}

\usepackage{multirow}
\usepackage{multicol}
\usepackage{graphicx}
\usepackage{tablefootnote}
\usepackage{threeparttable}
\usepackage{varwidth}
\usepackage{caption,newfloat}
\DeclareCaptionType{InfoBox}

\aclfinalcopy 

\title{Gendered Pronoun Resolution using BERT and an extractive question answering formulation}

\author{Rakesh Chada \\
  Jersey City, NJ \\
  \texttt{rakesh.chada@gmail.com} \\}

\date{}

\begin{document}
\maketitle
\begin{abstract}
    The resolution of ambiguous pronouns is a longstanding challenge in Natural Language Understanding. Recent studies have suggested gender bias among state-of-the-art coreference resolution systems. As an example, Google AI Language team recently released a gender-balanced dataset and showed that performance of these coreference resolvers is significantly limited on the dataset. In this paper, we propose\footnote{Source Code is available at \url{https://github.com/rakeshchada/corefqa}} an extractive question answering (QA) formulation of pronoun resolution task that overcomes this limitation and shows much lower gender bias (0.99) on their dataset. This system uses fine-tuned representations from the pre-trained BERT model and outperforms the existing baseline by a significant margin (22.2\% absolute improvement in F1 score) without using any hand-engineered features. This QA framework is equally performant even without the knowledge of the candidate antecedents of the pronoun. An ensemble of QA and BERT-based multiple choice and sequence classification models further improves the F1 (23.3\% absolute improvement upon the baseline). This ensemble model was submitted to the shared task for the 1st ACL workshop on Gender Bias for Natural Language Processing. It ranked 9th on the final official leaderboard.
\end{abstract}

\begin{table*}[!ht]
  \centering
  \renewcommand{\arraystretch}{1.2}
  \begin{tabular}{|p{5cm}|c|c|c|c|c|c|c|c|}
    \hline
    \multirow{2}{5cm}{\textbf{Number of examples}} & \multicolumn{4}{c|}{\textbf{Stage 1}} & \multicolumn{4}{c|}{\textbf{Stage 2}}\\
    \cline{2-9}
    & \textbf{T} & \textbf{A} & \textbf{B} & \textbf{N} & \textbf{T} & \textbf{A} & \textbf{B} & \textbf{N}\\
    \hline
    5-Fold Dev (80-20 split) & 2454 & 1105 & 1060 & 289 & 4454 & 1979 & 1985 & 490 \\ \hline
    Test & 2000 & 874 & 925 & 201 & 760 & 340 & 346 & 74  \\ \hline
  \end{tabular}
  \caption{\label{stats-table} Stage 1 and Stage 2 Dataset statistics.}
\end{table*}

\section{Introduction}
Coreference resolution is a task that aims to identify spans in a text that refer to the same entity. This is central to Natural Language Understanding. We focus on a specific aspect of the coreference resolution that caters to resolving ambiguous pronouns in English. Recent studies have shown that state-of-the-art coreference resolution systems exhibit gender bias~\cite{webster2018gap}~\cite{rudinger-EtAl:2018:N18}~\cite{zhao-etal-2018-gender}. ~\cite{webster2018gap} released a dataset that contained an equal number of male and female examples to encourage gender-fair modeling on the pronoun resolution task. A shared task for this dataset was then published on Kaggle\footnote{\url{https://www.kaggle.com/c/gendered-pronoun-resolution}}. The task involves classifying a specific ambiguous pronoun in a given Wikipedia passage as coreferring with one of the  three classes: first candidate antecedent (hereby referred to as \textbf{A}), second candidate antecedent (hereby referred to as \textbf{B}) or neither of them (hereby referred to as \textbf{N}). The authors show that even the best of the baselines such as~\cite{clark-manning-2015-entity},~\cite{wiseman-etal-2016-learning},~\cite{lee-etal-2017-end} achieve an F1 score of just 66.9\% on this dataset. The limited number of annotated labels available in this unbiased setting makes the modeling a challenging task. To that end, we propose an extractive question answering formulation of the task that leverages BERT~\cite{devlin2018bert} pre-trained representations and significantly improves (22.2\% absolute improvement in F1 score) upon the best baseline~\cite{webster2018gap}. In this formulation, the task is similar to a SQUAD~\cite{DBLP:conf/emnlp/RajpurkarZLL16} style question answering (QA) problem where the question is the context window (neighboring words) surrounding the pronoun to be resolved and the answer is the antecedent of the pronoun. The answer is contained in the provided Wikipedia passage. The intuition behind using the pronoun's context window as a question is that it allows the model to rightly identify the pronoun to be resolved as there can be multiple tokens that match the given pronoun in a passage. There has been previous work that cast the coreference resolution as a Question Answering problem~\cite{pmlr-v48-kumar16}. But the questions used in their approach take the form ``Who does ``she'' refer to?''. This would necessitate including additional information such as an indicator vector to identify the exact pronoun to be resolved when there are multiple of them in a given passage. Furthermore, their approach doesn't impose that the answer should be contained within the passage or the question text. ~\cite{mccann2018natural} model the pronoun resolution task of the Winograd schema challenge~\cite{levesque2012winograd} as a question answering problem by including the candidate antecedents as part of the question. An unique feature of the question answering framework (referred to as CorefQA) we propose is that it doesn't require the knowledge of the candidate antecedents in order to produce an answer for the pronoun resolution task. The model ``learns'', from training on the QA version of the shared task dataset, the specific task of extracting the appropriate antecedent of the pronoun given just the Wikipedia passage and the pronoun's context window. We also demonstrate other modeling variants for the shared task that use the knowledge of the candidate antecedents \textbf{A} and \textbf{B}. The first variant (CorefQAExt) is an extension of the CorefQA model that uses its predictions to produce probabilities over \textbf{A}, \textbf{B} and \textbf{N}. The second variant (CorefMulti) takes the formulation of a SWAG~\cite{zellers2018swag} style multiple choice classification and the final variant (CorefSeq) takes the standard sequence classification formulation. An ensemble of CorefQAExt, CorefMulti and CorefSeq models shows further performance gains (23.3\% absolute improvement in F1 score).

\section{Data}

The dataset used for this shared task is the GAP dataset~\cite{webster2018gap} where each row contains a Wikipedia text snippet, the corresponding page's URL, the pronoun to be resolved, the two candidate antecedents (\textbf{A} and \textbf{B}) of the pronoun, the text offsets corresponding to \textbf{A}, \textbf{B}, pronoun and boolean flags indicating the pronoun's coreference with A and B. The Kaggle competition for this shared task was conducted in two stages. Table~\ref{stats-table} shows the aggregate statistics for each stage. The 5-Fold Dev row represents the number of examples used for 5-fold stratified cross validation done based on the gender of the pronoun. This could lead to different distributions of A and B during the training of each fold. We chose to do so because we wanted to retain the perfect balance between male and female representations during training and thereby minimize the bias from the data. The columns \textbf{T}, \textbf{A}, \textbf{B} and \textbf{N} refer to the total number of examples, the number of examples where the pronoun's antecedent is A, B and neither respectively. We should note that for the question answering model, we exclude all the “neither” examples from the training data as we don’t have an exact answer. While this seems destructive, the model doesn't need, by design, an explicit supervision on the "neither" examples to predict an antecedent that's neither A nor B. The male and female pronoun examples are equally represented (50-50 split) in the development, validation and test datasets - with the exception of stage 2 test dataset. The stage 2 test dataset has 377 male and 383 female examples. We use lower-cased BERT word-piece tokenizer for preprocessing. This comes with a pre-built vocabulary of size 30522.

\section{System Description}

The final model used for submission is an ensemble of the question answering (CorefQAExt), multiple choice (CorefMulti) and sequence classification (CorefSeq) models. We describe each of these models in the following sections. We chose the pytorch-pretrained-bert\footnote{\url{https://github.com/huggingface/pytorch-pretrained-BERT}} library to implement all models. The source code is available at \url{https://github.com/rakeshchada/corefqa}

\begin{figure*}[!ht]
  \includegraphics[width=\linewidth]{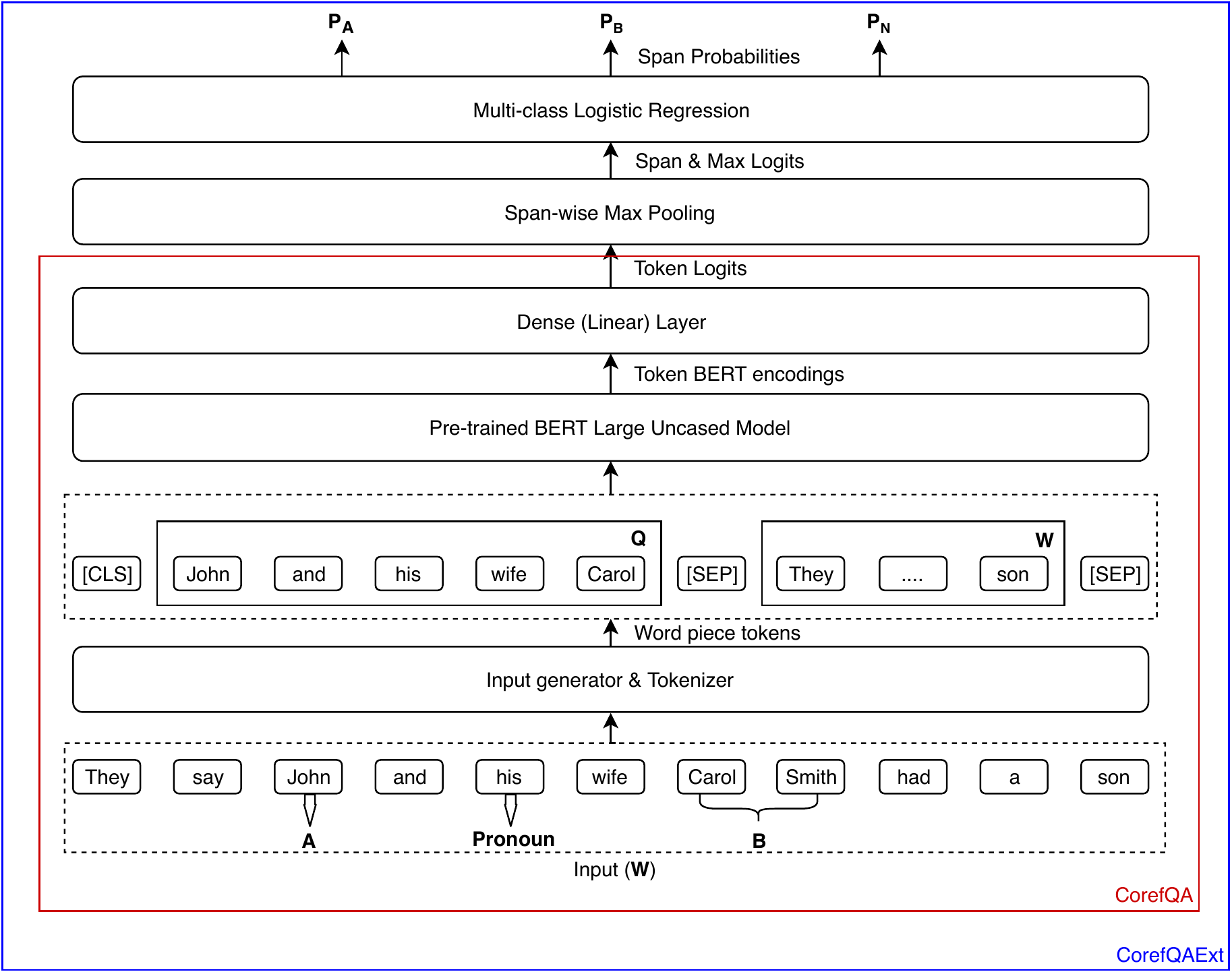}
  \caption{Architecture of CorefQA, CorefQAExt models}
  \label{fig:gapqa}
\end{figure*}

\subsection{Question Answering System (CorefQA, CorefQAExt)}
\subsubsection{Inputs and Architecture}

The architecture of this system is shown in Figure \ref{fig:gapqa}. The input \textbf{I} to the system can be represented as \textbf{I} = ``[CLS] \textbf{Q} [SEP] \textbf{W} [SEP]'' where \textbf{Q} represents the question text, \textbf{W} represents the Wikipedia passage text and [CLS], [SEP] are the delimiter tokens used in the BERT model. The question text \textbf{Q} is the pronoun context window of up to 5 words. The context window is the pronoun itself and its two neighboring words to the left and right. So, if \textbf{W} is \textit{``They say John and his wife Carol had a son''}, then \textbf{Q} would be \textit{``John and his wife Carol''} assuming ``\textit{his}'' is the pronoun to be resolved. In the case where there are less than two words on a given side, we just use the words available within the window - so these cases would lead to the window with less than 5 words. The text at this point is still un-tokenized so the ``words'' are just space separated tokens in a given text. The answer text is either \textbf{A}'s or \textbf{B}'s name (``neither'' cases have been initially filtered). The rest of the architecture until the Span-wise Max Pooling layer follows the standard SQUAD formulation in~\cite{devlin2018bert}. It's worth noting that the architecture until this point (before the Span-wise Max Pooling layer) doesn't use candidate antecedents' \textbf{A} and \textbf{B} text or offset information. The output at this intermediate layer (Dense Layer) contains two sets of logits: start and end logits for each token. These can then be used to extract the maximum scoring span as an answer as demonstrated in~\cite{devlin2018bert}. We refer to the architecture until the Span-wise Max Pooling Layer as CorefQA.

\subsubsection{Probability Estimation}

The shared task requires the output to be probabilities over the given \textbf{A}, \textbf{B} and \textbf{N} spans. So, we implement a mechanism that combines Span-wise Max Pooling and Logistic Regression to extract probabilities from start and end logits obtained in the previous step. Since we have access to offsets of \textbf{A} and \textbf{B}, we simply extract span logits corresponding to those offsets. Span logits are calculated by taking the maximum value of each of the individual token logits in a span. This gives us four values that represent maximum logits for the start and end of A and B spans. We also calculate maximum start and end logits over the entire sequence. These six logits are then fed as input features to a multi-class logistic regression. The output of this classifier then gives us the desired probabilities $\textbf{P}_A, \textbf{P}_B \& \textbf{P}_N$. We refer to this end-to-end architecture (from input layer to the Multi-class Logistic Regression layer) as CorefQAExt.

\subsubsection{Training \& Hyperparameters}

We use Adam optimizer with learning rate of 1e-5, $\beta_1$=0.9, $\beta_2$=0.999, L2 weight decay of 0.01, learning rate warmup over the first 10\% of total training steps, and linear decay of the learning rate. The maximum sequence length is set to 300 and batch size of 12 is used during training. We use BERT Large Uncased pre-trained model for initializing the weights of BERT layers. This model has 24 layers with each producing a 1024 dimensional hidden representation. The whole system is trained in an end-to-end fashion. We fine-tune the last 12 BERT Encoder layers (layer 13 to layer 24) and freeze layers 1 to 12 - meaning the parameters of those layers aren't updated during training. This leads to total trainable parameters in the order of 150 million. We didn't use any dropout. The hyperparameter C for the logistic regression is set to 0.1. This model was trained for 2 epochs on a NVIDIA K80 GPU. The training with the 5-fold cross validation finished in about 30 minutes. The average of the predictions of each fold on the test dataset is used as the final prediction. We had experimented with different choices for each of these hyperparameters - such as freezing or unfreezing more layers, choosing different learning rates, different batch sizes - but these numbers gave us the best results. Another hyperparameter the model was sensitive to was the context window size. Lower window sizes gave us better results with 5 being the ideal size.

\begin{table*}[!ht]
\begin{threeparttable}
  \centering
  \renewcommand{\arraystretch}{1.2}
  \begin{tabular}{|p{3cm}|c|c|c|c|c|c|c|c|c|c|}
    \hline
    \multirow{2}{3cm}{\textbf{Model}} & \multicolumn{5}{c|}{\textbf{Stage 1}} & \multicolumn{5}{c|}{\textbf{Stage 2}}\\
    \cline{2-11}
    & \textbf{M\tnote{*}} & \textbf{F\tnote{*}} & \textbf{B\tnote{*}} & \textbf{O\tnote{*}} & \textbf{L\tnote{*}} & \textbf{M\tnote{*}} & \textbf{F\tnote{*}} & \textbf{B\tnote{*}} & \textbf{O\tnote{*}} & \textbf{L\tnote{*}}\\
    \hline
    CorefQA & 88.8 & 87.8 & 0.99 & 88.3 & N/A\tnote{\#} & 93.2 & 91.3 & \textbf{1.0} & 92.2 & N/A\tnote{\#} \\ \hline
    CorefQAExt & \textbf{91.1} & 87.1 & 0.95 & 89.1 & 0.38 & 93.7 & 94.6 & \textbf{1.0} & \textbf{94.2} & 0.22 \\ \hline
    CorefMulti & 87.9 & 87.4 & \textbf{0.99} & 87.6 & 0.40 & 92.8 & 92.3 & 0.99 & 92.7 & 0.24 \\ \hline
    CorefSeq & 88.7 & 86.4 & 0.97 & 87.6 & 0.38 & 90.9 & 88.9 & 0.98 & 89.9 & 0.29  \\ \hline
    Full Ensemble & 90.9 & \textbf{89.5} & 0.98 & \textbf{90.2} & \textbf{0.32} & \textbf{94.1} & \textbf{94.0} & 1.0 & 94.0 & \textbf{0.20}  \\ \hline \hline
    QAMul Ensemble\tnote{+} & 91.1 & 88.4 & 0.97 & 89.7 & 0.35 & 93.9 & \textbf{94.3} & 1.0 & 94.1 & \textbf{0.19}  \\ \hline
  \end{tabular}
  \begin{tablenotes}\footnotesize
\item[*] L = Log-Loss, O = Overall F1, M = Male F1, F = Female F1, B = Bias (F/M)
\item[\#] N/A = Not Applicable
\item[+] Post competition Stage 2 deadline
\end{tablenotes}
  \caption{\label{results-table} Stage 1 and Stage 2 Test Results. \textbf{Bold} indicates best performance.}
  \end{threeparttable}
\end{table*}

\subsection{Multiple Choice classification (CorefMulti)}
Here, we formulate the task as a SWAG~\cite{zellers2018swag} style multiple choice problem among \textbf{A}, \textbf{B} and \textbf{N} classes.

\subsubsection{Inputs and Architecture}
For each example, we construct four input sequences, which each contain the concatenation of the the two sequences \textbf{S1} and \textbf{S2}. \textbf{S1} is a concatenation of the given Wikipedia passage with an additional sentence of the form ``\textbf{P} is '' where \textbf{P} is the text of the pronoun in question. So, for a passage that ends with the sentence \textit{``They say John and his wife Carol had a son''}, the sequence \textbf{S1} would be \textit{``They say John and his wife Carol had a son. his is ''} assuming \textit{``his''} is the pronoun to be resolved. The sequence \textbf{S2} is one of \textbf{A}'s name, \textbf{B}'s name or the word \textit{``neither''} if the pronoun in the example doesn't co-refer with \textbf{A} and \textbf{B}. Once we represent the inputs in this fashion, the rest of the architecture follows the design of BERT based SWAG task architecture discussed in~\cite{devlin2018bert}.

\subsubsection{Training \& Hyperparameters}
We use a batch size of 4 for training, initialize the BERT layers with the weights from the BERT Large Uncased pre-trained model and maintain the rest of the hyperparameters the same as the ones used for CorefQAExt model. Layers 12 to 24 of the BERT Encoder are fine-tuned and the rest of the layers are frozen. We use 5-fold cross validation with test prediction averaging from each fold. This model took about 100 minutes to run on Stage 1 data on a NVIDIA K80 GPU.

\subsection{Sequence classification (CorefSeq)}

This involves framing the problem as a standard sequence classification task. 

\subsubsection{Inputs and Architecture}

The input is the given Wikipedia passage without any additional augmentation. The sequence features are extracted by concatenating token embeddings corresponding to the A, B and the pronoun spans. These span embeddings are calculated by concatenating token embeddings of the start token, end token and the result of an element-wise multiplication of start and end token embeddings. The token embeddings are the output of the last encoder layer of the (fine-tuned) BERT. These features are then fed to a single hidden layer feed-forward neural network with a ReLU activation. This hidden layer has 512 hidden units. A softmax layer at the output then provides the desired \textbf{A}, \textbf{B} and \textbf{N} probabilities.

\subsubsection{Training \& Hyperparameters}

A dropout of 0.1 is applied before the inputs are fed from the BERT's last encoder layer to the feed forward neural network. The model is trained for 30 epochs with a batch size of 10. Layers 12 to 24 of the BERT Encoder are fine-tuned and the rest of the layers are frozen. A learning rate of 1e-5 is used with a triangular learning rate scheduler~\cite{smith2017cyclical} whose steps per cycle is set to 100 times the length of training data. We use 5-fold cross validation with test prediction averaging from each fold. This model took 105 minutes to run on Stage 1 data on a NVIDIA K80 GPU.

\section{Results and discussion}
Table~\ref{results-table} shows the results of all models for Stage 1 and Stage 2. We calculate Log-Loss, Male F1, Female F1, Overall F1 score and Bias (Female F1 / Male F1) as metrics on the test data sets. As the results show, all individual models improve upon the baseline model by a significant margin with the CorefQAExt model showing the highest absolute improvement of 22.2\%. It is interesting to note that the CorefQA model\footnote{Sample predictions shown in the Supplemental Material Section \ref{sec:supplemental}} still improved upon the baseline by 21.4\% despite not using the knowledge of candidate antecedents \textbf{A} and \textbf{B}. Infact, it slightly outperforms, on the Overall Stage 1 F1 score, both CorefMulti and CorefSeq models that explicitly encode the knowledge of \textbf{A} and \textbf{B}. A few input/output samples of the CorefQA model are shown in the Supplemental Section \ref{sec:supplemental}. It is worth noticing that this model (correctly) selects, most of the time, the spans corresponding to named entities as answers even though that constraint wasn't explicitly encoded in its design. The CorefQA model doesn't produce probabilities over \textbf{A}, \textbf{B} and \textbf{N} classes as that information isn't available to the model. Hence, we report Log-loss as ``N/A'' in Table~\ref{results-table}. The probabilities from the CorefQAExt, CorefMulti and CorefSeq are averaged to obtain the ensemble model’s probabilities. This ensemble model, with an Overall F1 score of 90.2, improves upon the baseline by 23.3 percentage points. This model ranked ninth on the final leaderboard of the Kaggle competition. The CorefMulti model seemed most robust to bias (0.99). The ensemble model had the best log loss in stage 2 even though the CorefQAExt model had the best Overall F1 score. This might be a reflection of the issues with probability calibration. Another explanation of this might be just the smaller stage 2 data size as compared to stage 1. Finally, although the CorefSeq model doesn't individually outperform other models, we get a better ensemble performance by including it rather than by excluding it.

\subsection{Freezing BERT weights}
We tried freezing all BERT layer weights for some of our initial experiments but hadn't seen much success - especially when we used the weights from the last encoder layer of the BERT. The Stage 1 Overall F1 score for the CorefQAExt model dropped down significantly to 63.6\% in this setting. This improved to 72.1\% if we used layer 18 weights. We also tried concatenating the last four encoder layer outputs of BERT. This resulted in an slightly better Overall F1 score of 74.4\% for Stage 1. So, the performance seemed to be sensitive to the choice of the encoder layer outputs. However, from the preliminary experiments, there seemed to be a big gap of about 15\% on the Overall F1 when compared to the fine-tuned model. A more principled \& thorough analysis of this phenomena makes an important future area of work.

\subsection{Post Stage 2 deadline Results}
After the competition had finished, we experimented with a few model variations on the final stage 2 test dataset that gave us interesting insights. Firstly, we tried excluding each model from the full ensemble. We noticed that we obtained a better Log Loss of 0.195 when we excluded CorefSeq. This model is listed as QAMul Ensemble in Table~\ref{results-table}. We carried another experiment where we trained the CorefQAExt using the cased version of the BERT model. An ensembling of the uncased version with this cased version delivered further performance gains (3\% absolute F1 improvement upon uncased CorefQAExt). Then, we tried ensembling the cased and uncased versions of all the three individual models - CorefQAExt, CorefMulti and CorefSeq on stage 2 test data. This resulted in an overall F1 score of 94.7\% , Male F1 of 94.8\%, Female F1 of 94.6\%, bias of 1.0 and a log loss of 0.197.

\subsection{Failed Experiments}
\begin{enumerate}
\item We tried fine-tuning the BERT model in an unsupervised manner by training a language model on the texts extracted from the Wikipedia pages corresponding to the URLs provided in the dataset. The idea behind this one was to see if we can get better BERT layer representations by tuning them to the shared task's dataset. However, this is a computationally expensive step to run and we didn't see promising gains from initial runs. We  hypothesize that this may be due to the fact that BERT representations were originally obtained by training on Wikipedia as one of the sources. So, fine-tuning on the task's dataset which is also from Wikipedia might not have added an extra signal.
\item For the CorefMulti model, we tried adding to the token embedding vector, an additional entity embedding vector that encodes the word-piece token level info of whether it belongs to one of \textbf{A}, \textbf{B} or \textbf{P}. We hypothesized this should help the model focus its attention on the relevant entities to the coreference task. But we weren't able to make a successful use of these embeddings to improve the model performance within the competition deadline. However, this is a promising future direction. 
\item For the CorefQAExt model, we appended the title extracted from the provided wikipedia page's URL into the input token sequence to evaluate if the page URL provides useful signal to the model. This made the performance slightly worse.
\end{enumerate}

\section{Conclusion}
We proposed an extractive question answering (QA) formulation of the pronoun resolution task that uses BERT fine-tuning and shows strong performance on the gender-balanced dataset. We have shown that this system can also effectively extract the antecedent of the pronoun without using the knowledge of candidate antecedents. We demonstrated three other formulations of the task that uses this knowledge. The ensemble of all these models obtained further gains (Table \ref{results-table}). This work showed that the pre-trained BERT representations provide a strong signal for the coreference resolution task. Furthermore, thanks to training on the gender-balanced dataset, this modeling framework was able to generate unbiased predictions despite using pre-trained representations. An important future work would be to analyze the gains obtained from BERT representations in more detail and perhaps compare it with alternate contextual token representations and fine-tuning mechanisms~\cite{peters-etal-2018-deep}~\cite{howard-ruder-2018-universal}. We also would like to apply our techniques to the Winograd schema challenge~\cite{levesque2012winograd}, the Definite Pronoun Resolution dataset~\cite{rahman2012resolving}, the Winogender schema dataset~\cite{rudinger-EtAl:2018:N18} and explore extensions to other languages perhaps using the CoNLL 2012 shared task dataset~\cite{pradhan2012conll}.

\section*{Acknowledgments}

We thank the GeBNLP committee reviewers for comments on the work and thank Thomas Wolf, Prashant Jayannavar, Hema Priya Darshini, Aquila Khanam for helpful feedback on the draft. We also thank the Google AI Language team for the Kaggle competition and the team at Hugging Face Inc. for the ``pytorch-pretrained-bert'' library.

\bibliography{acl2019}
\bibliographystyle{acl_natbib}

\appendix
\section{Supplemental Material}
\label{sec:supplemental}
This section lists a few example input/outputs of the CorefQA model that predicts answers to the gendered pronoun resolution task using just the Context and the Question (without the knowledge of the candidate antecedents \textbf{A} and \textbf{B}).

\begin{center}

\fbox{%
    \begin{varwidth}{0.45\textwidth}%
    \textbf{Context}: ``Alice (19), Kathleen Mary (12), Gertrude (10) and Mabel (7). In the 1901 census Allen was living at Fox Lane in Leyland with his 2nd wife Margaret (Whittle), daughter of James Whittle, a coachman, \& Ann Mills, whom he had married in 1900. She was some 18 years his junior.'' \\
    \textbf{Question}: ``1900. \textbf{She} was some''\\
    \textbf{Predicted Answer (Correct)}: ``Margaret (Whittle)''\\
    \end{varwidth}%
}\par
  \captionof{InfoBox}{CorefQA Prediction Sample 1\label{pred1}}
 
 \vspace{0.8cm}

\fbox{%
    \begin{varwidth}{0.45\textwidth}%
    \textbf{Context}: ``He then announced that CMU will celebrate Pausch's impact on the world by building and naming after Pausch a raised pedestrian bridge to connect CMU's new Computer Science building and the Center for the Arts, symbolizing the way Pausch linked those two disciplines. Brown University professor Andries van Dam followed Pausch's last lecture with a tearful and impassioned speech praising him for his courage and leadership, calling him a role model.'' \\
    \textbf{Question}: ``speech praising \textbf{him} for his''\\
    \textbf{Predicted Answer (Correct)}: ``Pausch''\\
    \end{varwidth}%
}\par
  \captionof{InfoBox}{CorefQA Prediction Sample 2\label{pred2}}
  
   \vspace{0.8cm}

\fbox{%
    \begin{varwidth}{0.45\textwidth}%
    \textbf{Context}: ``Walter S. Sheffer (August 7, 1918 - July 14, 2002) was an American photographer and teacher, born in Youngsville, Pennsylvania. He moved to Milwaukee, Wisconsin in 1945 to work at the studio of John Platz, Milwaukee's main society photographer. When Platz retired, Sheffer inherited his clientele and was able to establish his own ``look'' and very successful portrait studio by 1953.'' \\
    \textbf{Question}: ``Sheffer inherited \textbf{his} clientele and''\\
    \textbf{Predicted Answer (Wrong)}: ``Sheffer'\\
    \end{varwidth}%
}\par
  \captionof{InfoBox}{CorefQA Prediction Sample 3\label{pred3}}
  
   \vspace{0.8cm}

\fbox{%
    \begin{varwidth}{0.45\textwidth}%
    \textbf{Context}: ``I would never write a book about the bad parts. I would mostly revel in the fantastic parts, of which there were so many.'' In early 2007, reports surfaced concerning Lindsay Lohan's interest in buying the rights to Nicks' life story and developing a motion picture in which she planned to play her.'' \\
    \textbf{Question}: ``in which \textbf{she} planned to''\\
    \textbf{Predicted Answer (Wrong)}: ``Lindsay Lohan's interest in buying the rights to Nicks''\\
    \end{varwidth}%
}\par
  \captionof{InfoBox}{CorefQA Prediction Sample 4. The model wrongly predicts a bigger span as an answer.\label{pred4}}
  
  \vspace{0.8cm}
  
  \fbox{%
    \begin{varwidth}{0.45\textwidth}%
    \textbf{Context}: ``The president of SAG -- future United States President Ronald Reagan -- also known to the FBI as Confidential Informant ``T-10'', testified before the committee but never publicly named names. Instead, according to an FBI memorandum in 1947: ``T-10 advised Special Agent (name deleted) that he has been made a member of a committee headed by Mayer, the purpose of which is allegedly is to `purge' the motion-picture industry of Communist party members, which committee was an outgrowth of the Thomas committee hearings in Washington and subsequent meetings ....'' \\
    \textbf{Question}: ``) that \textbf{he} has been''\\
    \textbf{Predicted Answer (Correct)}: ``Special Agent''\\
    \end{varwidth}%
}\par
  \captionof{InfoBox}{CorefQA Prediction Sample 5 \label{pred5}}
  
    \vspace{0.8cm}

    \fbox{%
    \begin{varwidth}{0.45\textwidth}%
    \textbf{Context}: ``Emily Thorn Vanderbilt (1852--1946) was a member of the prominent United States Vanderbilt family. The second daughter of William Henry Vanderbilt (1821--1885) and Maria Louisa Kissam (1821--1896), Emily Thorn Vanderbilt was named after her aunt, Emily Almira (Vanderbilt) Thorn, daughter of dynasty founder Cornelius Vanderbilt.'' \\
    \textbf{Question}: `named after \textbf{her} aunt,''\\
    \textbf{Predicted Answer (Correct)}: ``Emily Thorn Vanderbilt''\\
    \end{varwidth}%
}\par
  \captionof{InfoBox}{CorefQA Prediction Sample 6 \label{pred6}}
  
  \end{center}
 
\end{document}